\documentclass[sigconf]{acmart}
\usepackage[ruled]{algorithm2e} 
\usepackage{tikz}
\usetikzlibrary{arrows,automata,positioning} 
\AtBeginDocument{%
  \providecommand\BibTeX{{%
    \normalfont B\kern-0.5em{\scshape i\kern-0.25em b}\kern-0.8em\TeX}}}

\setcopyright{acmcopyright}
\copyrightyear{2018}
\acmYear{2018}
\acmDOI{XXXXXXX.XXXXXXX}

\acmConference[X]{Y}
\acmPrice{15.00}
\acmISBN{978-1-4503-XXXX-X/18/06}

\begin{document}

\title{Computer vision based vehicle tracking as a complementary and scalable approach to RFID tagging}

\author{Pranav Kant Gaur}
\authornote{Both authors contributed equally to this research.}
\email{pranav@barc.gov.in}
\affiliation{%
  \institution{Bhabha Atomic Research Centre}
  \city{Mumbai}
  \state{Maharashtra}
  \country{India}
  \postcode{400085}
}

\author{Abhilash Bhardwaj}
\authornotemark[1]
\email{abhilashb@barc.gov.in}
\affiliation{%
  \institution{Bhabha Atomic Research Centre}
  \city{Mumbai}
  \state{Maharashtra}
  \country{India}
  \postcode{400085}
}

\author{Pritam Shete}
\email{ppshete@barc.gov.in}
\affiliation{%
  \institution{Bhabha Atomic Research Centre}
  \city{Mumbai}
  \state{Maharashtra}
  \country{India}
  \postcode{400085}
}
\author{Mohini Laghate}
\email{mlaghate@barc.gov.in}
\affiliation{%
  \institution{Bhabha Atomic Research Centre}
  \city{Mumbai}
  \state{Maharashtra}
  \country{India}
  \postcode{400085}
}
\author{Dinesh M Sarode}
\email{dinesh@barc.gov.in}
\affiliation{%
  \institution{Bhabha Atomic Research Centre}
  \city{Mumbai}
  \state{Maharashtra}
  \country{India}
  \postcode{400085}
}

\renewcommand{\shortauthors}{Gaur and Bhardwaj, et al.}

\begin{abstract}
  Logging of incoming/outgoing vehicles serves as a piece of critical information for root-cause analysis to combat security breach incidents in various sensitive organizations. RFID tagging hampers the scalability of vehicle tracking solutions on both logistics as well as technical fronts. For instance, requiring each incoming vehicle(departmental or private) to be RFID tagged is a severe constraint and coupling video analytics with RFID to detect abnormal vehicle movement is non-trivial. We leverage publicly available implementations of computer vision algorithms to develop an \textit{interpretable} vehicle tracking algorithm using finite-state machine formalism. The state-machine consumes input from the cascaded object detection and optical character recognition(OCR) models for state transitions. We evaluated the proposed method on 75 video clips of 285 vehicles from our system deployment site. We observed that the detection rate is most affected by the speed and the type of vehicle. The highest detection rate is achieved when the vehicle movement is restricted to follow a movement restrictions(SOP) at the checkpoint similar to RFID tagging. We further analyzed 700 vehicle tracking predictions on live-data and identified that the majority of vehicle number prediction errors are due to illegible-text, image-blur, text occlusion and out-of-vocab letters in vehicle numbers. Towards system deployment and performance enhancement, we expect our ongoing system monitoring to provide evidences to establish a higher vehicle-throughput SOP at the security checkpoint as well as to drive the fine-tuning of the deployed computer-vision models and the state-machine to establish the proposed approach as a promising alternative to RFID-tagging.
\end{abstract}



%
\maketitle

\section{Introduction}
Logging of incoming/outgoing vehicles serves as a piece of critical information for root-cause analysis to combat security breach incidents in various sensitive organizations. Subject to the severity of various internal maintenance and expansion activities, such organizations witness a large inflow of both internal(or official) as well as external(or private) vehicles, thereby, rendering the manual tracking of vehicle movement, cumbersome, resource-intensive, time-consuming, and most importantly ineffective in providing rapid and reliable inputs. RFID vehicle tagging alleviates speed and accuracy limitations but limits the applicability to departmental vehicles, as private vehicles with temporary entry permits cannot be tagged without introducing further delays in the process. Further, in practice, coupling video solutions--for instance to track vehicle direction-- with RFID technology has been tricky, as multiple incoming vehicles can be detected in arbitrary order, and capturing video/photographic evidence sometimes generates data without the RFID reported vehicle, thereby rendering it useless. On the other hand, computer vision algorithms, owing to their rapid advancement in the public domain, provide access to state-of-art implementations for problems like object detection, and optical character recognition(OCR) which can be leveraged to build a reliable, scalable and a complementary solution RFID tagging. 

\textbf{\textit{Related work}}: 
Distinct vehicle counting and tracking are the active research areas pertaining to smart transportation, surveillance and security. Researchers involved in these domains have followed diverse methodologies to tackle this problem. Broadly, these solutions are based on frame differencing, counting by detection and density estimation. With advancement of deep learning and computer vision, researchers have also applied deep learning in field of detection and tracking. In \cite{Yang2020} author presented a detection based tracking method, where vehicles are detected using YOLO and then a lightweight feature extraction network is used to extract discriminative features. Author has also discussed the robustness of this method in case of external interferences like occlusion. In \cite{Hao2014} author has proposed a novel particle filter algorithm for vehicle tracking. Author has also proposed block symmetry in observation model to improve results in background with similar colors, under partial occlusion. In \cite{1706793} author trained a tracking specific model with automatic feature selection. In this work, discriminative features are constructed from pairs of image patches. In \cite{https://doi.org/10.48550/arxiv.2007.16198} authors have conducted experiments with different detection and tracking models. They have deployed and tested deep learning solutions like CenterNet and Deep SORT, Detectron2 and Deep SORT, YOLOv4 and  Deep SORT for detection and tracking. In \cite{8705792} a vehicle tracking algorithm is proposed for smart traffic networks. In this author presents track by detection paradigm, where vehicle detection is followed by extended IoU tracker. Extended IoU tracker is fused with historical tracking information to improve the robustness of tracking.

\textbf{\textit{{Contribution of this work}}}:
We have demonstrated that a performant and \textit{interpretable}\cite{https://doi.org/10.48550/arxiv.2202.02540} baseline approach for vehicle tracking could be developed over publicly available object detection and OCR models using finite-state formalism without resorting to deep-learning based object tracking methods. We highlight the effect of vehicle classification bias on vehicle detection rate as a limitation of the reported method as well as the ways to combat it. Similar to the spirit of vehicle tracking, we reaffirm a known fact that the off-the-shelf OCR models are robust enough to reduce the need for a complicated prediction-filtering logic over vehicle-number predictions generated across multiple image-frames of a vehicle.

\section{Problem formulation}
A vehicle is considered as \textit{logged} if an entry as shown in Table-\ref{tab:veh-log} is generated once the corresponding vehicles arrive at the security checkpoint.
\begin{table}[H]
  \begin{tabular}{cccl}
    \toprule
    Vehicle number & Vehicle type & Vehicle timestamp \\
    \midrule
    MH03CS0071 & Car & June 7 2022 10:40:00 GMT \\
    KA06N9659 & Bus & June 12 2022 12:23:00 GMT \\
  \bottomrule
\end{tabular}
  \caption{ANPR vehicle log}
  \label{tab:veh-log}
\end{table}

\subsection{Constraints}
Vehicles movement is bidirectional, Vehicle numbers may include \textit{special characters} like $\uparrow$ in addition to the regular vehicle number alphabets for India. Vehicle classes include: Car, Jeep, Bus and Truck, it is important to have uniform vehicle tracking and number prediction performance across all vehicle classes.  A very high vehicle number-prediction accuracy is desired because departmental vehicles are procured in bulk, and therefore, get assigned \textit{consecutive} vehicle numbers, one letter flip can map the prediction to another valid vehicle number.

\subsection{Metric}
The vehicle-detection rate\cite{uk-anpr} is the fraction of the vehicles detected by the system to the total number of observed vehicles for a specific duration. The vehicle number prediction from images can be treated as an applied case of Optical character recognition, therefore,  a negation of Word-error-rate\cite{4376991} or the Word accuracy(WA)--referred here as the vehicle-number prediction accuracy--has been used to assess the quality of vehicle number predictions. 

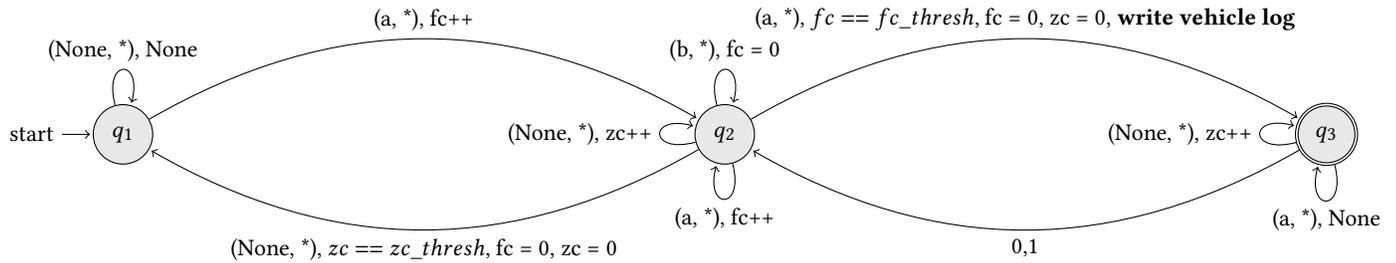
\begin{figure*}[t]
    \centering
\begin{tikzpicture}[shorten >=1pt,node distance=8cm,on grid,auto]
  \tikzstyle{every state}=[fill={rgb:black,1;white,10}]

    \node[state,initial]   (q_1)                    {$q_1$};
    \node[state] (q_2)  [right of=q_1]    {$q_2$};
    \node[state,accepting]           (q_3)  [right of=q_2]    {$q_3$};
    
    \path[->]
    (q_1) edge [loop above] node {(None, *), None}   (   )
          edge [bend left]  node {(a, *), fc++}    (q_2)
    (q_2) edge [loop below] node {(a, *), fc++}   ( )
          edge [loop left] node {(None, *), zc++} ()
          edge [bend left]  node {(a, *), $fc == fc\_thresh$, fc = 0, zc = 0, \textbf{write vehicle log}}    (q_3)
          edge [loop above] node {(b, *), fc = 0}  ()
          edge [bend left]  node {(None, *), $zc == zc\_thresh$, fc = 0, zc = 0}  (q_1)
    (q_3) edge [bend left]  node {0,1}  (q_2)
          edge [loop left] node {(None, *), zc++} ()
          edge [loop below] node {(a, *), None} ();
     %

\end{tikzpicture}
    \caption{Vehicle tracking automaton}
    \label{fig:veh-track-auto}
\end{figure*}

\section{Methodology}
Vehicle tracking algorithm receives inputs from vehicle type, number-plate and vehicle-number prediction models. Algorithm \ref{algo: overview} summarises the overall approach:

\begin{algorithm}
\SetKwProg{trackVehicle}{Function \emph{trackVehicle}}{}{end}

\trackVehicle{}{
frame\_counter = 0\;
zero\_counter = 0\;
frame\_thresh = $\alpha$\;
zero\_thresh = $\beta$\;
last\_state = 0\;
\ForAll{image $img$ in image\_stream}{
     lp\_bbox, lp\_pred, v\_class = pfrs\_process($img$, pred\_confidence)\;
     next\_state = pers\_process(lp\_bbox, lp\_pred, last\_state, frame\_counter, zero\_counter)\;
     last\_state = next\_state\;
     }}
\caption{Proposed vehicle tracking algorithm}
\label{algo: overview}
\end{algorithm}

The algorithm takes each image as input and passes it to the function $pfrs\_process$ which encapsulates the joint vehicle-detection(with class prediction) followed by vehicle-number prediction. The vehicle tracking algorithm, denoted here as $pers\_process$ initializes in \textit{no vehicle detected} state and receives number-plate bounding box, vehicle class and number-plate text predictions from $pfrs\_process$ to updates its internal state with corresponding actions, which when transitioning from \textit{vehicle-detected in current image} to \textit{a unique vehicle detected(across last few frames)} state implies writing a new vehicle entry(ref. Table-\ref{tab:veh-log}) in the vehicle-log database.

\subsection{Joint Vehicle and Number-plate detection}
The objective of vehicle and number-plate detection stage is to filter number-plate bounding box prediction for downstream consumption by number-plate prediction model. $pfrs\_process$ receives image from an image-source(a video, webcam-feed or live-stream) and invokes YOLOv5 model\cite{glenn_jocher_2022_6222936} for vehicle and number-plate detection. In each image-frame, multiple-vehicles and number-plates may get detected, therefore, we apply the following 3-stage prediction filtering criterion(in order): 1. All predictions must have the prediction confidence beyond a threshold. 2. A number-plate prediction is passed downstream only if it lies within a vehicle bounding-box. 3. Vehicle bounding box predictions are then filtered out and the remaining number-plate bounding box predictions are sent downstream.

\subsection{Number-plate recognition}
The objective of number-plate recognition model is to generate vehicle number prediction corresponding to the cropped image received from the joint vehicle and number-plate detection model. Considering the generality of OCR data-sets(\cite{Jaderberg16}) and the variety of Indian number-plates\cite{https://doi.org/10.48550/arxiv.2207.06657} both in terms of number-plate background and text-font, we have used an open-source OCR model, the PaddleOCR\cite{https://doi.org/10.48550/arxiv.2009.09941} for vehicle number recognition. Unlike YOLOv5, we did not require fine-tuning PaddleOCR for majority of vehicle numbers, but defense vehicle numbers(with $\uparrow$) will require model fine-tuning. 

\subsection{Vehicle tracking algorithm}
The vehicle-tracking algorithm has been modelled as a finite-state automaton as shown in fig. \ref{fig:veh-track-auto}. The automaton starts in $q_1$, which represent \textit{no vehicle detected} condition and updates its state and takes corresponding actions for each image-frame received from the video-source. Inputs for the automaton are number-plate bounding box and number-plate text predictions received from $pfrs\_process$. State $q_2$ represents \textit{vehicle detected in current image-frame} condition. System stays in $q_2$ until it has not received sufficient number of \textit{similar} vehicle bounding box predictions, represented by the state variable $frame\_counter$(or $fc$). For all state-transitions except $q_2$ to $q_3$, actions of automaton involve updating its state-variables like $frame\_counter$ and $zero\_counter$(or $zc$) based on heuristically derived $frame\_threshold$(or $f\_thresh$) and $zero\_threshold$(or $z\_thresh$), which represent, the minimum votes -- in terms of the number of successive predictions and no-predictions, respectively -- required to make a state-transition. $zero\_counter$ accumulates the number of no-detection image-frames in any state, in order to prevent frequent transitions to $q_1$. For $q_2$ to $q_3$, the automaton writes the vehicle log for the vehicle just detected to the database.

\subsection{Standard operating procedure}
A SOP defines spatial and temporal restrictions on the movement of vehicles at the security checkpoint. The vehicle is expected to stop within a rectangular region defined as the ROI for a minimum duration(0.6 seconds). The ROI constraint assists the vehicle number prediction by providing it with an optimal trade-off between image-quality and perspective distortion, whereas the minimum time-duration assists vehicle tracking algorithm by providing it sufficient number of frames to register a track as valid. 

\section{Experiments}
\subsection{Ground-truth generation process}
We have evaluated our method on both test-videos as well as live-stream. The test video clips have been collected covering all times for an entire week. Out of 75 test-clips, 65 clips include vehicles following SOP, whereas 10 videos have a mix of vehicles under SOP and rush-hours(>80 vehicles/hour). For each video, we have manually recorded the entire vehicle-log, resulting in a database of 285 vehicles across entire week sampled at specific times of the days. Achieving accurate vehicle number-predictions under rainfall condition has been a challenging task, therefore, we have monitored live predictions of our system over 700 vehicle movement instances during the month of July in Mumbai. 

\subsection{Implementation}
We have fine-tuned the PyTorch implementation of YOLOv5 base-model trained over COCO dataset\cite{10.1007/978-3-319-10602-1_48} for vehicle and number-plate detection over 150 images spread over vehicle classes. Specifically, we annotated each input image with vehicle bounding boxes, vehicle-classes(car, bus, jeep and truck) and number-plate coordinates. The model was fine-tuned for 250 epochs with batch-size 16 and 640X640 input size. 

For number-plate recognition, we have used the publicly available PP-OCRv2 model for English text \cite{https://doi.org/10.48550/arxiv.2009.09941}. The vehicle-tracking algorithm is implemented in Python. The system runs over a virtual machine with Quadro-M4000 GPU in our on-premise cloud service and accesses live feed from IP camera installed at security checkpoint.

\subsection{Context for experiments}
The proposed system is targeted towards the gradual automation of the manual vehicle-logging practice at the security checkpoint of our organization. While the duplicate detection of a vehicle is tolerable for system-acceptance in initial stages, a \textit{missed} vehicle serves as a red-flag. The vehicle detection rate is therefore, the most crucial aspect of the system. Our goal has been to establish a data-driven vehicle-movement policy(or SOP) to be able to ensure close to 100\% detection rate in subsequent iterations.  
In table-\ref{tab:sop-det-rate}, we report the sensitivity of the detection-rate of the proposed method to vehicle movement constraints, 
\begin{table}[H]
  \caption{Effect of vehicle movement on detection rate}
  \label{tab:sop-det-rate}
  \begin{tabular}{ccl}
    \toprule
    Vehicle movement & Detection rate & \# of instances\\
    \midrule
    Vehicle following SOP & 89.6 & 155\\
    Vehicle violating SOP & 42.3 & 130 \\
  \bottomrule
\end{tabular}
\end{table}

Once, the SOP is established as a way to achieve reasonable performance, we experimented with the \textit{fairness} of our method with respect to its detection-rate performance across majority and minority vehicle-classes. From the data-collection activity to train, validate and monitor our system, we observed that around 98\% of vehicles at the security checkpoint are either cars(17.5\%) or jeeps(80.1 \%), whereas, the trucks and bus classes represent only the remaining 2\% of the traffic. The system, however, should have uniform detection rate across all vehicle classes. Given the dependence of detection bounding box filtering phase on vehicle-classes(ref. sec. 3.1 ), we measured the detection rate of our system across these vehicle classes.  Table-\ref{tab:veh-class-det-rate} provides the quantitative evidence for our observation that there is vehicle-class bias in our proposed method.  

\begin{table}[H]
  \caption{Effect of vehicle-class on detection rate}
  \label{tab:veh-class-det-rate}
  \begin{tabular}{cccl}
    \toprule
    Vehicle class(under SOP) & Detection rate(\%) & \# of instances\\
    \midrule
    Car, Jeep &  94 & 139 \\
    Truck, Bus & 25  & 8\\
  \bottomrule
\end{tabular}
\end{table}

We assessed the robustness of vehicle number prediction over rainfall scenario for the system running over live-stream from the deployment site. Table-\ref{tab:veh-no-pred} enlists the impact of various issues we observed across 700 vehicle detection instances once the system was evaluated in rainy season under SOP movement restrictions. \begin{table}[H]
  \caption{Effect of image-quality, input-vocab on Vehicle number prediction}
  \label{tab:veh-no-pred}
  \begin{tabular}{cccl}
    \toprule
    Issue & Drop in WA(\%) & \# of instances(out of 700)\\
    \midrule
    Illegible text & 6 & 42\\
    Motion blur &  2.5 & 18 \\
    Text occlusion & 2.5 & 18\\
    $\uparrow$ vehicle numbers & 1.2 & 9 \\
  \bottomrule
\end{tabular}
\end{table}

\section{Discussion}
Table-\ref{tab:sop-det-rate} established that there is a strong correlation between placing restrictions over vehicle movement at the security checkpoint and the detection rate. The proposed method, however, still misses vehicles under SOP conditions. The primary reasons based on debugging those cases revealed that the vehicle-class detection model generates predictions for rarely occurring vehicle-classes with lower confidence resulting in filtering of those predictions, due to which the tracking algorithm does not receive such predictions as input for state transitions. We can address this either by removing such bias from the vehicle-detection model by employing upsampling, data-augmentation and weighted loss-function approaches\cite{10.1145/3191442.3191458, Johnson2019} for model-training or we can decouple the prediction filtering process from vehicle-class predictions. We would prefer model-training approach as it will also help us address the issues reported in Table-\ref{tab:veh-class-det-rate}.

In Table-\ref{tab:veh-no-pred}, we identified the major factors affecting the vehicle number prediction performance. Given that the most impactful issue was observed to be illegibility of the text, it provided us with the evidence that publicly available OCRs now match human-performance. Text occlusion are cases where the number-plate text is covered with dirt, rust or partially visible due to an occlusion from another vehicle or security operator. Incorrect predictions due to motion blur result not only from the mismatch between vehicle-speed and shutter-speed of IP camera, but also due to a naive prediction selection strategy currently employed which selects the last prediction before $q_2$ to $q_3$ transition(ref. Figure-\ref{fig:veh-track-auto}) as the representative vehicle number. Performance drop due to motion-blur, therefore, could be addressed first by using more sophisticated strategies, like vehicle number string clustering, fine-tuning PaddleOCR in addition to increasing the frame-threshold for vehicle-tracking algorithm, thereby allowing vehicle to come to a halt. Most importantly, the evidence from this monitoring exercise have helped us prioritize fine-tuning existing OCR models over training number-plate recognition models from scratch\cite{https://doi.org/10.48550/arxiv.1808.08410}. In terms of the effort involved to monitor predictions on daily basis, the reported exercise took 12 man-hours, after which we started noticing patterns in the vehicle number prediction issues which could be monitored in algorithmic fashion, giving us sufficient motivation to \textit{automate} the monitoring process in near future \cite{https://doi.org/10.48550/arxiv.2007.06299, Rukat2020}.

\section{Conclusions}
We have demonstrated that under the standard operating conditions of security checkpoints, like toll booths, it is feasible to achieve high vehicle-logging accuracy using an \textit{interpretable} vehicle-tracking algorithm leveraging existing open-domain implementations of object detection and OCR models without resorting to deep-learning based object-tracking solutions. Experiments revealed that the automaton is sensitive to its vehicle-class input which could be addressed by removing corresponding bias in the underlying vehicle detection model. Our work is in progress along developing experimental evidences for establishing a high vehicle-throughput SOP for peak-hour traffic at the security checkpoint, addressing model bias against rarely encountered vehicle classes, as well as automation of our system-monitoring activity. 

\bibliographystyle{ACM-Reference-Format}
\bibliography{CODS}

\end{document}